\documentclass{article}
\usepackage{graphicx} 

\title{Empowering Distributed Solutions in Renewable Energy Systems and Grid Optimization}
\author{Mohammad Mohammadi, Ali Mohammadi}
\date{September 2023}

\begin{document}

\maketitle

\section{Abstract}
This study delves into the shift from centralized to decentralized approaches in the electricity industry, with a particular focus on how machine learning (ML) advancements play a crucial role in empowering renewable energy sources and improving grid management. ML models have become increasingly important in predicting renewable energy generation and consumption, utilizing various techniques like artificial neural networks, support vector machines, and decision trees. Furthermore, data preprocessing methods, such as data splitting, normalization, decomposition, and discretization, are employed to enhance prediction accuracy.

The incorporation of big data and ML into smart grids offers several advantages, including heightened energy efficiency, more effective responses to demand, and better integration of renewable energy sources. Nevertheless, challenges like handling large data volumes, ensuring cybersecurity, and obtaining specialized expertise must be addressed. The research investigates various ML applications within the realms of solar energy, wind energy, and electric distribution and storage, illustrating their potential to optimize energy systems. To sum up, this research demonstrates the evolving landscape of the electricity sector as it shifts from centralized to decentralized solutions through the application of ML innovations and distributed decision-making, ultimately shaping a more efficient and sustainable energy future.

\section{Introduction}

Due to the rapid advancement of global industrialization, it is acknowledged that excessive use of fossil fuels will not just speed up the depletion of these resources but will also harm the environment. These consequences will lead to elevated health risks and the looming threat of worldwide climate change \cite{lai2020survey}. Traditional fossil fuel-based energy sources are under pressure due to uncertainty about their role in global warming \cite{rangel2021machine}. Alongside fossil fuels and nuclear power, renewable energy is currently the fastest-growing energy sector. The increasing focus on recent studies can be attributed to the growing popularity of sustainable and environmentally friendly renewable energy sources such as solar power, wind energy, hydroelectric power, biomass, waves, tides, and geothermal energy due to their minimal environmental impact. A primary challenge for renewable energy in the near future is energy supply. This refers to the integration of renewable energy resources into existing or future energy supply frameworks \cite{wang2019review}. Consequently, focus is shifting to alternative energy sources that need to substantially increase their contributions to global energy production in the upcoming years, as societies move away from oil and coal. Clean and renewable energies play a pivotal role here, but their implementation presents unique challenges that necessitate the development of new technologies \cite{rangel2021machine}.

The growing energy demand has made the adoption of renewable sources inevitable. Over time, many power companies have established renewable energy facilities worldwide to offer economical and environmentally friendly energy \cite{missaoui2014managing}. Renewable sources like wind and solar power come with advantages such as lower delivery costs and reduced emissions. However, traditional grid energy storage designs are becoming less practical. Occasional large-scale blackouts have emphasized the need for an enhanced decision-making process that relies on timely and precise data regarding dynamic events, operating conditions, and sudden power changes \cite{mostafa2022renewable}.

Rifkin \cite{rifkin2011third} has defined the energy internet as an innovative energy utilization system that merges renewable energy sources, decentralized power stations, hydrogen energy, storage technologies, and electric vehicles with Internet technologies. The author has outlined four attributes of the energy internet: reliance on renewable energy, support for access to large-scale generation and storage systems, facilitation of energy sharing, and promotion of electric transportation. In contrast to fossil fuel-based power plants, managing renewable energy necessitates more advanced control of power, equilibrium, and production capacity, attainable through smart grids \cite{rathor2020energy}. These grids combine traditional power networks with advanced Information Technology (IT) and communication networks to distribute electricity with greater efficiency and reliability, while also reducing costs and environmental impacts \cite{yan2012survey}. 

The shift of the electricity grid towards the smart grid is facilitated by the Internet of Things (IoT), an interconnected network of sensing and actuating devices that enables seamless sharing of information across platforms through a unified framework. This framework creates a cohesive operating view to support innovative applications. This is achieved through pervasive and continuous sensing, data analysis, and information representation, with cloud computing serving as the overarching infrastructure. Each of these interconnected devices possesses its own embedded computing system, enabling identification and interaction with other devices, and often necessitates the use of advanced data analysis and machine learning techniques \cite{rifkin2011third}.

These technologies bring about various advantages, including enhanced energy efficiency, improved demand response, and better integration of renewable energy sources. However, it's important to acknowledge potential drawbacks, such as the vulnerability to cyber-attacks and the complexity inherent in machine learning algorithms. Highlighting the need to address security concerns within the smart grid, it outlines potential security threats and corresponding countermeasures.

The applications of IoT extend across different layers of the smart grid. In the Generation layer, it facilitates monitoring energy generation, controlling units, tracking gas emissions and pollution, predicting power consumption, and managing distributed power plants and microgrids. In the Transmission layer, it enables monitoring and control of transmission lines and substations, along with the production management of transmission towers. The Distribution layer benefits from IoT by automating distribution processes, managing and safeguarding equipment, and handling faults. At the Consumption layer, IoT is applied to smart homes and appliances, intelligent charging and discharging of electric vehicles \cite{en16176379}, power load control, and multi-network management \cite{hossain2019application}.

This evolving landscape requires new strategies for energy production management, forecasting, and prediction. A significant challenge lies in the fact that renewable energies are not entirely controllable, as their generation heavily depends on environmental factors like wind, cloud cover, and rainfall. A potential solution to address these issues involves expert systems based on Machine Learning (ML) algorithms, capable of handling the nonlinearities and complex modeling prevalent in current systems \cite{perera2014machine}.

Prediction holds significant importance in the management of renewable energy, particularly in critical sectors like wind and solar power. Prediction of resources is a widespread practice, encompassing various scales, ranging from large-scale facilities like wind and solar farms to smaller setups such as micro-grids with limited generation resources \cite{jung2017optimal}. Effective renewable energy forecasting serves as a crucial tool for addressing uncertainties, thereby supporting the planning, management, and operation of electrical power and energy systems \cite{frias2017assessing}.

Despite the inherent advantages of sustainable energy sources, several technical factors impede their broader adoption and higher integration within the power grid. The primary challenge is the intermittence of these resources, which presently hinders their more substantial involvement in the energy mix. The irregular availability of these resources underscores the pressing need for the development of precise prediction systems aimed at estimating the power generated by renewable sources \cite{salcedo2018feature}. Achieving accuracy in renewable energy forecasting remains complex due to the unpredictable and chaotic nature of renewable energy data. This unpredictability stems from various factors such as weather patterns, cloud cover, and wind speed, all of which influence the amount of energy produced by renewable sources \cite{hodge2018combined}. The uncertainties introduced by these factors can introduce instability into the power system and diminish its overall stability margin \cite{wang2019review}.

Probabilistic forecasting models play a pivotal role in the realm of renewable energy prediction, offering a crucial advantage by furnishing quantitative measures of uncertainty associated with renewable energy data. The conventional deterministic point forecasts often fall short in capturing the inherent variability of renewable energy data. In contrast, probabilistic forecasts emerge as a valuable tool for enhancing planning, management, and operation within electric energy systems. This is achieved by assigning probabilities to different prediction outcomes, thus yielding a comprehensive perspective.

Probabilistic forecasting encompasses two overarching categories: parametric and nonparametric methods, each with or without distributional assumptions. Parametric methods generally operate under the presumption that renewable energy time series data adheres to a predefined distribution, which could be Gaussian, beta, or Gamma distributions, among others. Once this distribution is established, diverse statistical techniques are employed to assess its parameters. Methods such as auto-regression models, maximum likelihood estimation, and rapid Bayesian approaches are often utilized to determine these parameters, thereby enabling the creation of a comprehensive probability set that underpins a probabilistic prediction \cite{wang2019review}.

\section{Empowering Renewable Energy Through Machine Learning Innovations}

Machine-learning techniques aim to uncover relationships between input and output data, either with or without explicit mathematical formulations. Once these machine-learning models are effectively trained using a dataset, decision-makers can generate accurate forecasted output values by inputting forecast data into the trained models. \cite{learning2007review}
While some studies have concentrated on forecasting renewable energy through the use of a single machine-learning model \cite{amasyali2018review}, the diverse nature of datasets, time intervals, prediction spans, configurations, and performance metrics make it challenging to enhance forecast performance through a single model. Consequently, to boost prediction accuracy, certain studies have developed hybrid machine-learning models or comprehensive prediction approaches tailored for renewable energy forecasts. Recently, support vector machines and deep learning methods have gained substantial popularity in the realm of machine learning \cite{wang2019review}.

Jung-Pin Lai \cite{lai2020survey} covers a range of machine-learning models, encompassing artificial neural networks, support vector machines, decision trees, and random forests. The paper outlines the merits and demerits of each model and furnishes instances of their applications in renewable energy predictions. For instance, artificial neural networks are widely adopted in predicting renewable energy due to their capacity to comprehend intricate connections between input and output variables. Nevertheless, they demand extensive training data and can be computationally intensive. Support vector machines are another prevalent machine-learning model utilized in renewable energy forecasts. Their prowess lies in handling high-dimensional data, suitable for both regression and classification tasks. However, they are sensitive to the selection of kernel functions and necessitate careful hyperparameter tuning. Decision trees and random forests also find frequent application in renewable energy prediction. While decision trees are straightforward and interpretable, they might encounter issues of overfitting and instability. Random forests, an extension of decision trees, mitigate overfitting and enhance predictive precision. Yet, they can be computationally demanding and require ample training data.

The data pre-processing phase holds a critical role in machine learning, significantly elevating the efficiency of machine-learning performance \cite{learning2007review}. Numerous prevalent data pre-processing techniques are employed in machine-learning models for predicting renewable energy. These techniques are implemented to ready the data for analysis and enhance the accuracy of predictions. One commonly used pre-processing technique is data splitting, involving the division of data into training and testing sets. This separation enables the evaluation of machine-learning model performance on unseen data, guarding against overfitting. Normalization is another technique, entailing the scaling of data to a consistent range. This practice ensures that all variables carry equal significance in the analysis, curbing bias towards variables with larger values. Decomposition is a pre-processing approach applied in renewable-energy predictions, involving the breakdown of time series data into its constituent trend, seasonal, and residual elements. This strategy aims to filter out noise and uncover underlying patterns within the data. Discretization is yet another pre-processing technique, converting continuous variables into discrete categories. This simplification aids in analysis and reduces data dimensionality. Further pre-processing techniques in renewable-energy predictions encompass feature selection, imputation of missing values, encoding categorical features, and standardization. The selection of a pre-processing technique hinges on the specific context and the type of renewable-energy source under consideration \cite{lai2020survey}.

The integration of big data and machine learning within the smart grid holds several potential benefits, including improved energy efficiency, more effective demand response, enhanced incorporation of renewable energy sources, and refined load forecasting. Moreover, these technologies have the capability to lower operational expenses and heighten system reliability. Nonetheless, it is imperative to acknowledge potential drawbacks as well \cite{hossain2019application}.

The researchers acknowledge that the dataset utilized in their study is relatively limited for comprehensive big data analytics. Nevertheless, they assert that the integration of cloud computing and real-time event analysis has established a fitting framework for big data analytics. Their suggestions for future investigations encompass the incorporation of larger datasets encompassing diverse renewable energy sources and demand patterns across multiple countries. Furthermore, they propose involving customers in data input and supplying information about their energy usage to enhance the precision of predictive models \cite{mostafa2022renewable}.

One challenge stems from the vast volume of data produced by the smart grid, which can prove challenging to manage and analyze, necessitating substantial computational resources. The risk of cyber-attacks and data breaches is also a critical concern, given the potential dire implications for both the power grid and its users. Furthermore, the utilization of machine learning algorithms can be intricate, demanding specialized expertise. This complexity might act as a hindrance to adoption for certain organizations. Finally, apprehensions revolve around the privacy of user data and the potential misuse of this information. These concerns underscore the need for careful implementation and robust security measures when leveraging these technologies within the smart grid framework \cite{hossain2019application}.
Several distinct instances highlight the application of big data science within the smart grid context. These instances encompass load forecasting \cite{coelho2017gpu}, demand response \cite{zhou2015demand} \cite{zhou2016big}, fault detection and diagnosis, detection of energy theft, determination of residential energy consumption \cite{zhou2016understanding}, and the integration of renewable energy \cite{bessa2015probabilistic}. Additionally, machine learning-driven algorithms have been devised to oversee power quality events, analyze user preferences, and manage the scheduling of residential loads \cite{hossain2019application}. 

Machine learning (ML) models have found widespread utility across various aspects of energy systems, particularly in forecasting electrical energy and renewable energy demand and consumption. These models offer pathways to enhance energy efficiency and reduce costs in the energy sector through diverse avenues. For instance, accurate energy consumption and demand predictions generated by ML models can be harnessed by building commissioning project managers, utility companies, and facility managers to implement energy-saving strategies. ML models also serve load forecasting, power generation prediction, power quality estimation, time series forecasting, wind speed projection, and power demand anticipation, among other applications \cite{mosavi2019state}. Moreover, the prediction of building energy consumption holds paramount importance in shaping decisions aimed at curbing energy usage and CO2 emissions. It aids in evaluating various building design alternatives, operational strategies for energy efficiency, and refining demand and supply management \cite{amasyali2018review}. However, predicting building energy consumption remains challenging due to the multitude of factors influencing it, including building attributes, installed equipment, outdoor weather conditions, and occupants' energy-use patterns \cite{kwok2011study}.

Deep learning is a type of machine learning that is capable of discovering the inherent nonlinear features and high-level invariant structures in data. This makes it particularly well-suited for forecasting renewable energy, which is characterized by intermittent and chaotic data. Deep learning algorithms are able to identify patterns and relationships in the data that may not be apparent using other machine learning techniques, which can lead to more accurate predictions. Deep learning stands as a promising avenue for enhancing renewable energy forecasting \cite{wang2019review}.
The ability to accurately predict energy demand and consumption equips energy companies to optimize their production and distribution operations, resulting in cost reductions and enhanced energy efficiency. Moreover, machine learning (ML) models demonstrate their proficiency in managing optimization tasks like storage planning, energy control, peak load management, dynamic energy pricing, cost minimization, and the estimation of battery charging requirements \cite{mosavi2019state}. 

The hybrid machine learning technique enhances the predictability of renewable energy consumption by harnessing the strengths of different models. In a study by Rasha \cite{abd2022renewable}, a hybrid approach was employed that integrates three fundamental models: Cat-Boost, SVR, and MLP. Cat-Boost, a gradient boosting algorithm, is recognized for its adeptness in managing categorical features and delivering high accuracy and efficiency. SVR, a support vector regression technique, offers transparent computation during dataset prediction and estimation. MLP, a multilayer perceptron and a type of artificial neural network, is employed for supervised learning.
In this hybrid model, historical load data from a previous time period serves as the training dataset, and a specific algorithm is selected to train the data network. The network's structure is meticulously designed to elevate the performance and predictive capacity of renewable energy consumption, particularly in challenging scenarios. Validation is performed using a separate test dataset, with diverse error metrics employed to analyze and assess the outcomes.

Cat-Boost's competence in handling categorical features, coupled with SVR's straightforward computation during dataset prediction, and MLP's potential to enhance performance and predictive accuracy of renewable energy consumption, contribute to the hybrid model's effectiveness. This hybrid approach exhibits superior accuracy and efficiency compared to other methods, positioning it as a promising strategy for forecasting renewable energy consumption.
To evaluate the proposed system's effectiveness, a range of assessment metrics is employed. These metrics encompass mean absolute error (MAE), mean squared error (MSE), root mean squared error (RMSE), and the coefficient of determination (R2). These metrics collectively offer insights into the precision and efficacy of the hybrid machine learning model proposed for predicting renewable energy consumption \cite{abd2022renewable}.

Nevertheless, the integration of ML models into energy systems is not devoid of challenges. One key hurdle pertains to the availability and quality of data, significantly influencing the accuracy and dependability of ML forecasts. The inherent complexity of energy systems introduces another layer of difficulty, complicating the development of accurate and efficient ML models. Additionally, certain ML models lack transparency, posing challenges in comprehending their predictive logic – a concern for stakeholders. The computational demands of specific ML models, especially in real-time scenarios, also impose limitations. Despite these obstacles, the growing body of literature detailing ML's contributions to energy systems underscores the substantial potential of these models in this arena \cite{mosavi2019state}.

\section{Revolutionizing Renewable Energy and Grid Management through Machine Learning}

The implications for the future of renewable energy across diverse industries and households are poised to be substantial \cite{abd2022renewable}. In the machine learning approach, historical load data from a previous time period is selected as the training dataset. This data is used to formulate an appropriate network structure, followed by the application of a specific training algorithm to fine-tune the network's performance, with the ultimate goal of meeting predefined accuracy benchmarks \cite{khan2020machine}. The influence of machine learning on energy systems spans various dimensions. In this study, particular attention is directed towards solar energy, wind power, and electric distribution and storage. Among these, wind power production has garnered significant attention, as evidenced by the numerous prediction models that have been proposed \cite{costa2008review}. This is attributed to wind power being one of the most advanced and widely utilized renewable sources. Nonetheless, there is an observable rise in the development of models for other renewable sources such as solar \cite{yadav2014solar} and marine energy \cite{cuadra2016computational} in recent years.

\subsection{Unveiling ML Applications in Solar Energy}

As the integration of solar energy into the energy system continues to expand, the accurate prediction of solar power generation becomes increasingly crucial. This prediction plays a vital role in managing energy quality and enhancing the overall reliability of the system \cite{khan2020machine}.

In the domain of solar energy forecasting, Machine Learning (ML) techniques such as Deep Neural Networks, Support Vector Machines, and Random Forest models have been effectively utilized to enhance the precision of solar irradiance predictions. These methods offer advantages such as refined forecasts through their ability to capture nonlinear relationships, improved weather projections, and expedited forecasting. However, they necessitate sample datasets for training, longer training durations, and a broad spectrum of sensory input. The suitability of Support Vector Machines might be restricted in this context.

ML models, particularly Artificial Neural Networks, play a pivotal role in foreseeing and monitoring the performance of solar energy systems. These models act as substitutes to mechanistic models and facilitate system optimization. However, they can be influenced by decreased accuracy due to measurement errors.

In the realm of power output forecasting for solar energy systems, ML techniques like Artificial Neural Networks, Support Vector Machines, hybrid models, and regression trees have proven beneficial. These approaches tackle challenges such as instability and inaccuracies in measurements. Yet, they are better suited for short-term predictions and might face difficulties in handling abnormal weather conditions. The creation of hybrid models for such systems can be intricate.

For tasks like optimizing material composition and designing components of solar energy systems to enhance performance, ML techniques like Deep Neural Networks, Genetic Algorithms, and Random Forest models excel. They can identify optimal designs from a multitude of possibilities and propose innovative models rooted in existing knowledge. However, modeling the design of such systems can be complex, and acquiring experimental data can be costly and demanding.\cite{rangel2021machine}

\subsection{Unveiling ML Applications in Wind Energy System}

Leveraging Machine Learning (ML) for predicting wind power, both in the short and long term, involves the application of various models including Regression models, deep neural networks, support vector machines, and decision trees. These methodologies offer benefits like predictions closely matching power curves and the flexibility to adapt to extreme weather situations. However, they do encounter challenges such as potential inaccuracies in forecasting during atypical weather patterns and the complexities arising from intricate variable relationships.

Wind energy systems are exposed to vulnerabilities arising from external factors and the operation of moving components, particularly blade susceptibility to delamination. ML techniques like k-nearest neighborhood, neural networks, decision trees, and support vector machines are employed to anticipate maintenance requirements. These methods yield advantages such as minimizing on-site interventions through ML-driven wind farm monitoring. Nevertheless, they face challenges including the need for a high sampling frequency, a substantial number of variables for effective failure detection, and the wide array of potential faults.

For localizing wind speed predictions, uncovering feature correlations, and projecting wind speed patterns, ML approaches like Neural networks, decision trees, and support vector regression models are harnessed. These techniques offer benefits such as augmenting weather predictions with historical data. However, constraints arise from their applicability to short-term predictions, potential prediction failures during unusual weather conditions, and the requirement for multiple features to attain accurate training.\cite{rangel2021machine}

\subsection{Unveiling ML Applications in Electric Distribution and Storage}

In the realm of the smart grid, Machine Learning (ML) plays a pivotal role in facilitating decisions related to pricing, consumption, production, fraud detection, and security measures. Techniques such as Q-learning, deep neural networks, support vector machines, and reinforcement learning are harnessed to extract insights from smart grid data. These methodologies offer advantages like real-time data integration and organization for informed decision-making, ensuring the secure operation of the smart grid. However, they grapple with limitations including the need for substantial data volumes, diverse heterogeneous information sources, and susceptibility to minor errors during unforeseen events.

Energy load forecasting, which accounts for weather conditions, is enhanced through methods like Deep neural networks and support vector machines. These techniques provide benefits like accurate predictions of short- and long-term loads using ML models. Nonetheless, they require significant data volumes, time-intensive testing and validation procedures, and extensive CPU resources for training.

ML proves instrumental in predicting stable compounds for novel material design, especially in the quest for alternative battery materials. Support vector regression, generative models, artificial neural networks, and hybrid models combined with DFT models are employed. These approaches yield advantages such as identifying promising components like electrodes and electrolytes. However, they face challenges in terms of data volume requirements and often rely on simulated data to complement experimental data.
For predicting properties of battery materials for component and structure projection in batteries, techniques such as Artificial neural networks, deep neural networks, and support vector regression models are utilized. These methods offer benefits like real-time optimization with reduced computational demands. Yet, the scarcity of experimental data and the potential need to integrate ML techniques with first-principles models (hybrid models) pose limitations.

Optimal management of battery charging, a complex task reliant on substantial battery data, is aided by ML methods like Extreme learning machines, artificial neural networks, and reinforcement learning. These approaches offer advantages such as real-time optimization with decreased computational requirements. Nevertheless, they encounter challenges tied to limited experimental data and the potential necessity for hybrid models that combine ML with first-principles models.\cite{rangel2021machine}

\section{Safeguarding the Future: Navigating Security Concerns in Smart Grid Technology}

Across every facet of the smart grid—generation, transmission, and distribution—there exists a substantial vulnerability to cyber-attacks, with several such attacks having already occurred. Consequently, data security emerges as a primary apprehension within the smart grid framework, prompting extensive efforts towards identifying cyber-security threats and establishing protective measures to counter them. Many of these defense strategies have harnessed machine learning techniques, as traditional methods often prove inadequate in this novel data-driven, non-linear context. Nonetheless, further exploration is necessary to formulate effective solutions for other security considerations like physical threats, network assaults, and encryption breaches. The refinement of communication systems, bolstered by enhanced protective measures, is also imperative. Therefore, security concerns hold a pivotal role in shaping the implementation of these technologies within the power grid \cite{hossain2019application}.

The paper \cite{su2021secure} introduces a novel approach to address the challenges of secure and efficient data sharing in smart grids using federated learning-enabled AIoT (Artificial Intelligence of Things). The proposed scheme focuses on enhancing the security, efficiency, and privacy concerns that hinder the implementation of AIoT services based on federated learning within smart grids. The primary objective of the scheme is to create a secure and efficient framework for sharing private energy data while maintaining data privacy and optimizing communication.

The scheme presents an innovative edge-cloud-assisted federated learning framework that aims to overcome issues related to low-quality shared models, non-IID (non-independent and identically distributed) data distributions, and unpredictable communication delays. By incorporating both edge and cloud resources, the framework facilitates communication-efficient and privacy-preserving sharing of energy data within smart grids. Additionally, the scheme introduces a mechanism for evaluating local data in the federated learning process, considering the heterogeneous and non-IID nature of user data.

 Federated Learning (FL) is a decentralized approach to machine learning where edge agents generate models based on their local data and contribute to a global model on a server. This process maintains data privacy to some extent since raw data is not directly shared with the server. Only a fraction of interested agents participate in the FL training process, simplifying the process and improving efficiency. The FL process involves multiple iterations of server-agent interaction to train the model and produce a smart model by learning from local agents. Initially, a global model is shared with all participants, and each agent performs local training epochs by dividing their data into batches and optimizing the model. The prepared local models are then shared with the server.
The FL process follows a series of three steps. Firstly, the FL server initializes a global model for a specific task and chooses a group of agents known as participants. This global model is then distributed to the participating agents. Secondly, each participant receives the global model and conducts on-device training using their local data while also benefiting from the global model. The participant generates a local model based on this training and shares it with the server. Finally, the FL server collects the local models from the participants and combines their model parameters to create an updated global model. A new subset of agents is selected by the server, and the updated global model is shared with them. This iterative process continues until the global model reaches the desired level of convergence.\cite{imteaj2022leveraging}

Furthermore, the scheme formulates the optimization problems and payoff functions for both users and Energy Service Providers (ESPs) under the context of the federated learning framework. To incentivize user participation and encourage high-quality contributions to the model, a two-layer deep reinforcement learning (DRL)-based incentive algorithm is devised.
Overall, the proposed scheme presents a comprehensive approach to addressing security, efficiency, and privacy concerns in energy data sharing within smart grids. By leveraging federated learning, edge-cloud collaboration, and innovative incentive mechanisms, the scheme holds the potential to enhance the overall performance and effectiveness of AIoT services in the context of smart grids. \cite{su2021secure}

\section{The role of distributed decision-making and information processing in the future energy systems}

The broad adoption of distributed energy resources (DERs), particularly renewable sources of decentralized generation, offers significant potential to greatly improve the efficiency of electricity distribution. However, as DERs become a substantial part of the total energy on the distribution network, inadequate integration processes could lead to imbalanced and unstable temporary behaviors. These behaviors could strain the existing infrastructure, possibly resulting in power outages and drops in voltage. In a future scenario of a smart grid, consumers equipped with renewable generation capabilities like solar panels and wind turbines could use predictive strategies to optimize how they consume energy. They would determine when to use, sell, or store the renewable energy they generate themselves. This active engagement with the electric grid and other consumers would replace the current passive energy consumption model.

Communication among distributed nodes (consumers) equipped with generation, storage, and consumption capabilities could establish a decentralized framework for decision-making and control. This approach would lead to improvements in both overall energy efficiency and cost reduction. However, to fully realize the potential benefits of the smart grid concept, a systematic set of design principles is necessary, along with a comprehensive protocol framework that facilitates interaction among the various entities in the grid. Furthermore, robust and computationally efficient control and optimization algorithms are crucial elements of this endeavor.

The Energy Management System (EMS) utilizes the optimal power flow (OPF) algorithm to optimize the performance of Distributed Energy Resources (DERs) within a microgrid. By analyzing power network data, the OPF algorithm determines optimal set points for DERs \cite{olivares2014trends}. Its primary objective is twofold: minimize the overall cost of the power system while adhering to technical and physical constraints of both the power network and the DERs. The speed of optimization varies based on specific scenarios. To achieve this, the EMS relies on diverse input data, including predictions for renewable generator output, local load, Battery Energy Storage Systems (BESSs) charge status, operational limits of dispatchable generators and BESSs, microgrid security and reliability thresholds, ancillary service requirements from the utility grid, and anticipated energy prices \cite{tushar2014three}.

After the OPF optimization, the determined optimal set points are relayed to the primary control systems of the DERs. These set points serve as references for active and reactive power, instructing the internal voltage and current control loops of the fast and dynamically efficient power electronic converters integrated within the DERs. This meticulous process ensures accurate tracking of the designated power values, leading to heightened stability and enhanced overall performance of the microgrid.

The interaction between utility companies and consumers, involving both energy generation and consumption aspects, plays a crucial role in the design of smart grids. This interaction differs from traditional power grid planning, which mainly focuses on matching generation with demand. By integrating considerations of generation and consumption, the scheduling of power and loads can be optimized more effectively. However, the nature of this utility-consumer interaction varies depending on the time scales of interaction periods and the different consumer units within the hierarchical structure.

For example, in a microgrid setup, smart homes act as consumer units at the microgrid level, while the microgrid controller serves as a consumer unit at the feeder level, which is one level above. To accommodate the diverse ways in which generation and consumption sides interact, a two-stage model for utility-consumer interaction has been suggested. This model consists of two phases: initial scheduling, which corresponds to long-term planning, and real-time scheduling, which pertains to short-term planning.\cite{li2011load}\cite{li2011auctioning}

The implementation of demand response (DR) strategies during the initial scheduling establishes a baseline operating point for nodes in the grid, characterized by consistent consumer load patterns. However, since these DR schemes rely on predictions of renewable generation over the scheduling period (e.g., 12 or 24 hours), they may not effectively handle the real-time fluctuations and intermittency in the power grid due to the inevitable differences between actual and predicted renewable generation.
To tackle this challenge and improve overall efficiency and stability, it's advantageous to introduce interactions at a finer time scale (short-term) between utility companies and consumers (both on the generation and consumption sides), nearly in real time. From the consumers' perspective, who often prioritize their own interests, the objective is to make optimal decisions that maximize cumulative profits or minimize expenses by utilizing their local distributed energy resources (DERs).

Given the relatively consistent load profiles established by DR schemes, consumers can choose to sell surplus renewable energy to the grid while storing the rest for future use. These decisions are guided by real-time pricing information, allowing consumers to dynamically respond to market conditions.

Designing a distributed decision scheme for selling excess energy within a microgrid entails addressing several important factors. Firstly, the scheme should be adaptable to changes in consumer behavior, accommodating the potential number of smart homes within the microgrid and their dynamic shifts between buying and selling modes. Secondly, the scheme must enable selling-mode smart homes to express their willingness to sell surplus energy units, ensuring that the units sold align with high willingness levels. To quantify this willingness, a measure needs to be established, considering potential variations across different smart homes and even within the same home due to changing energy unit availability. As a result, the microgrid controller must ensure that energy units sold consistently correspond to elevated willingness metrics across different consumers. Lastly, the scheme should be resilient against collusion among self-interested selling-mode smart homes, which might provide false willingness metrics for personal gain. By addressing these factors, a comprehensive distributed decision scheme can be developed to enhance the efficiency of energy sales within the microgrid.\cite{li2014distributed}

\bibliographystyle{ieeetr}

\bibliography{Ref}

\begin{thebibliography}{10}

\bibitem{lai2020survey}
J.-P. Lai, Y.-M. Chang, C.-H. Chen, and P.-F. Pai, ``A survey of machine
  learning models in renewable energy predictions,'' {\em Applied Sciences},
  vol.~10, no.~17, p.~5975, 2020.

\bibitem{rangel2021machine}
D.~Rangel-Martinez, K.~Nigam, and L.~A. Ricardez-Sandoval, ``Machine learning
  on sustainable energy: A review and outlook on renewable energy systems,
  catalysis, smart grid and energy storage,'' {\em Chemical Engineering
  Research and Design}, vol.~174, pp.~414--441, 2021.

\bibitem{wang2019review}
H.~Wang, Z.~Lei, X.~Zhang, B.~Zhou, and J.~Peng, ``A review of deep learning
  for renewable energy forecasting,'' {\em Energy Conversion and Management},
  vol.~198, p.~111799, 2019.

\bibitem{missaoui2014managing}
R.~Missaoui, H.~Joumaa, S.~Ploix, and S.~Bacha, ``Managing energy smart homes
  according to energy prices: analysis of a building energy management
  system,'' {\em Energy and Buildings}, vol.~71, pp.~155--167, 2014.

\bibitem{mostafa2022renewable}
N.~Mostafa, H.~S.~M. Ramadan, and O.~Elfarouk, ``Renewable energy management in
  smart grids by using big data analytics and machine learning,'' {\em Machine
  Learning with Applications}, vol.~9, p.~100363, 2022.

\bibitem{rifkin2011third}
J.~Rifkin, {\em The third industrial revolution: how lateral power is
  transforming energy, the economy, and the world}.
\newblock Macmillan, 2011.

\bibitem{rathor2020energy}
S.~K. Rathor and D.~Saxena, ``Energy management system for smart grid: An
  overview and key issues,'' {\em International Journal of Energy Research},
  vol.~44, no.~6, pp.~4067--4109, 2020.

\bibitem{yan2012survey}
Y.~Yan, Y.~Qian, H.~Sharif, and D.~Tipper, ``A survey on smart grid
  communication infrastructures: Motivations, requirements and challenges,''
  {\em IEEE communications surveys \& tutorials}, vol.~15, no.~1, pp.~5--20,
  2012.

\bibitem{en16176379}
M.~Mohammadi, J.~Thornburg, and J.~Mohammadi, ``Towards an energy future with
  ubiquitous electric vehicles: Barriers and opportunities,'' {\em Energies},
  vol.~16, no.~17, 2023.

\bibitem{hossain2019application}
E.~Hossain, I.~Khan, F.~Un-Noor, S.~S. Sikander, and M.~S.~H. Sunny,
  ``Application of big data and machine learning in smart grid, and associated
  security concerns: A review,'' {\em Ieee Access}, vol.~7, pp.~13960--13988,
  2019.

\bibitem{perera2014machine}
K.~S. Perera, Z.~Aung, and W.~L. Woon, ``Machine learning techniques for
  supporting renewable energy generation and integration: a survey,'' in {\em
  Data Analytics for Renewable Energy Integration: Second ECML PKDD Workshop,
  DARE 2014, Nancy, France, September 19, 2014, Revised Selected Papers 2},
  pp.~81--96, Springer, 2014.

\bibitem{jung2017optimal}
J.~Jung and M.~Villaran, ``Optimal planning and design of hybrid renewable
  energy systems for microgrids,'' {\em Renewable and Sustainable Energy
  Reviews}, vol.~75, pp.~180--191, 2017.

\bibitem{frias2017assessing}
L.~Fr{\'\i}as-Paredes, F.~Mallor, M.~Gast{\'o}n-Romeo, and T.~Le{\'o}n,
  ``Assessing energy forecasting inaccuracy by simultaneously considering
  temporal and absolute errors,'' {\em Energy Conversion and Management},
  vol.~142, pp.~533--546, 2017.

\bibitem{salcedo2018feature}
S.~Salcedo-Sanz, L.~Cornejo-Bueno, L.~Prieto, D.~Paredes, and
  R.~Garc{\'\i}a-Herrera, ``Feature selection in machine learning prediction
  systems for renewable energy applications,'' {\em Renewable and Sustainable
  Energy Reviews}, vol.~90, pp.~728--741, 2018.

\bibitem{hodge2018combined}
B.-M. Hodge, C.~B. Martinez-Anido, Q.~Wang, E.~Chartan, A.~Florita, and
  J.~Kiviluoma, ``The combined value of wind and solar power forecasting
  improvements and electricity storage,'' {\em Applied Energy}, vol.~214,
  pp.~1--15, 2018.

\bibitem{learning2007review}
S.~M. Learning, ``A review of classification techniques, sb kotsiantis,'' {\em
  Informatica}, vol.~31, pp.~249--268, 2007.

\bibitem{amasyali2018review}
K.~Amasyali and N.~M. El-Gohary, ``A review of data-driven building energy
  consumption prediction studies,'' {\em Renewable and Sustainable Energy
  Reviews}, vol.~81, pp.~1192--1205, 2018.

\bibitem{coelho2017gpu}
I.~M. Coelho, V.~N. Coelho, E.~J. d.~S. Luz, L.~S. Ochi, F.~G. Guimaraes, and
  E.~Rios, ``A gpu deep learning metaheuristic based model for time series
  forecasting,'' {\em Applied Energy}, vol.~201, pp.~412--418, 2017.

\bibitem{zhou2015demand}
K.~Zhou and S.~Yang, ``Demand side management in china: The context of
  china’s power industry reform,'' {\em Renewable and Sustainable Energy
  Reviews}, vol.~47, pp.~954--965, 2015.

\bibitem{zhou2016big}
K.~Zhou, C.~Fu, and S.~Yang, ``Big data driven smart energy management: From
  big data to big insights,'' {\em Renewable and sustainable energy reviews},
  vol.~56, pp.~215--225, 2016.

\bibitem{zhou2016understanding}
K.~Zhou and S.~Yang, ``Understanding household energy consumption behavior: The
  contribution of energy big data analytics,'' {\em Renewable and Sustainable
  Energy Reviews}, vol.~56, pp.~810--819, 2016.

\bibitem{bessa2015probabilistic}
R.~J. Bessa, A.~Trindade, C.~S. Silva, and V.~Miranda, ``Probabilistic solar
  power forecasting in smart grids using distributed information,'' {\em
  International Journal of Electrical Power \& Energy Systems}, vol.~72,
  pp.~16--23, 2015.

\bibitem{mosavi2019state}
A.~Mosavi, M.~Salimi, S.~Faizollahzadeh~Ardabili, T.~Rabczuk, S.~Shamshirband,
  and A.~R. Varkonyi-Koczy, ``State of the art of machine learning models in
  energy systems, a systematic review,'' {\em Energies}, vol.~12, no.~7,
  p.~1301, 2019.

\bibitem{kwok2011study}
S.~S. Kwok and E.~W. Lee, ``A study of the importance of occupancy to building
  cooling load in prediction by intelligent approach,'' {\em Energy Conversion
  and Management}, vol.~52, no.~7, pp.~2555--2564, 2011.

\bibitem{abd2022renewable}
R.~M. Abd El-Aziz, ``Renewable power source energy consumption by hybrid
  machine learning model,'' {\em Alexandria Engineering Journal}, vol.~61,
  no.~12, pp.~9447--9455, 2022.

\bibitem{khan2020machine}
P.~W. Khan, Y.-C. Byun, S.-J. Lee, D.-H. Kang, J.-Y. Kang, and H.-S. Park,
  ``Machine learning-based approach to predict energy consumption of renewable
  and nonrenewable power sources,'' {\em Energies}, vol.~13, no.~18, p.~4870,
  2020.

\bibitem{costa2008review}
A.~Costa, A.~Crespo, J.~Navarro, G.~Lizcano, H.~Madsen, and E.~Feitosa, ``A
  review on the young history of the wind power short-term prediction,'' {\em
  Renewable and Sustainable Energy Reviews}, vol.~12, no.~6, pp.~1725--1744,
  2008.

\bibitem{yadav2014solar}
A.~K. Yadav and S.~Chandel, ``Solar radiation prediction using artificial
  neural network techniques: A review,'' {\em Renewable and sustainable energy
  reviews}, vol.~33, pp.~772--781, 2014.

\bibitem{cuadra2016computational}
L.~Cuadra, S.~Salcedo-Sanz, J.~Nieto-Borge, E.~Alexandre, and
  G.~Rodr{\'\i}guez, ``Computational intelligence in wave energy: Comprehensive
  review and case study,'' {\em Renewable and Sustainable Energy Reviews},
  vol.~58, pp.~1223--1246, 2016.

\bibitem{su2021secure}
Z.~Su, Y.~Wang, T.~H. Luan, N.~Zhang, F.~Li, T.~Chen, and H.~Cao, ``Secure and
  efficient federated learning for smart grid with edge-cloud collaboration,''
  {\em IEEE Transactions on Industrial Informatics}, vol.~18, no.~2,
  pp.~1333--1344, 2021.

\bibitem{imteaj2022leveraging}
A.~Imteaj and M.~H. Amini, ``Leveraging asynchronous federated learning to
  predict customers financial distress,'' {\em Intelligent Systems with
  Applications}, vol.~14, p.~200064, 2022.

\bibitem{olivares2014trends}
D.~E. Olivares, A.~Mehrizi-Sani, A.~H. Etemadi, C.~A. Ca{\~n}izares,
  R.~Iravani, M.~Kazerani, A.~H. Hajimiragha, O.~Gomis-Bellmunt, M.~Saeedifard,
  R.~Palma-Behnke, {\em et~al.}, ``Trends in microgrid control,'' {\em IEEE
  Transactions on smart grid}, vol.~5, no.~4, pp.~1905--1919, 2014.

\bibitem{tushar2014three}
W.~Tushar, B.~Chai, C.~Yuen, D.~B. Smith, K.~L. Wood, Z.~Yang, and H.~V. Poor,
  ``Three-party energy management with distributed energy resources in smart
  grid,'' {\em IEEE Transactions on Industrial Electronics}, vol.~62, no.~4,
  pp.~2487--2498, 2014.

\bibitem{li2011load}
D.~Li, S.~K. Jayaweera, O.~Lavrova, and R.~Jordan, ``Load management for
  price-based demand response scheduling-a block scheduling model,'' {\em Proc.
  ICREPQ}, pp.~1--6, 2011.

\bibitem{li2011auctioning}
D.~Li, S.~K. Jayaweera, and A.~Naseri, ``Auctioning game based demand response
  scheduling in smart grid,'' in {\em 2011 IEEE online conference on green
  communications}, pp.~58--63, IEEE, 2011.

\bibitem{li2014distributed}
D.~Li and S.~K. Jayaweera, ``Distributed smart-home decision-making in a
  hierarchical interactive smart grid architecture,'' {\em IEEE Transactions on
  Parallel and Distributed Systems}, vol.~26, no.~1, pp.~75--84, 2014.

\end{thebibliography}

\end{document}